\documentclass[10pt,twocolumn,letterpaper]{article}

\usepackage{cvpr}              
\usepackage{colortbl}

\usepackage[T1]{fontenc}
\usepackage{amsfonts}

\definecolor{cvprblue}{rgb}{0.21,0.49,0.74}
\usepackage[pagebackref,breaklinks,colorlinks,allcolors=cvprblue]{hyperref}

\def\paperID{*****} 
\def\confName{CVPR}
\def\confYear{2025}

\title{VideoGLaMM \includegraphics[width=0.04\linewidth]{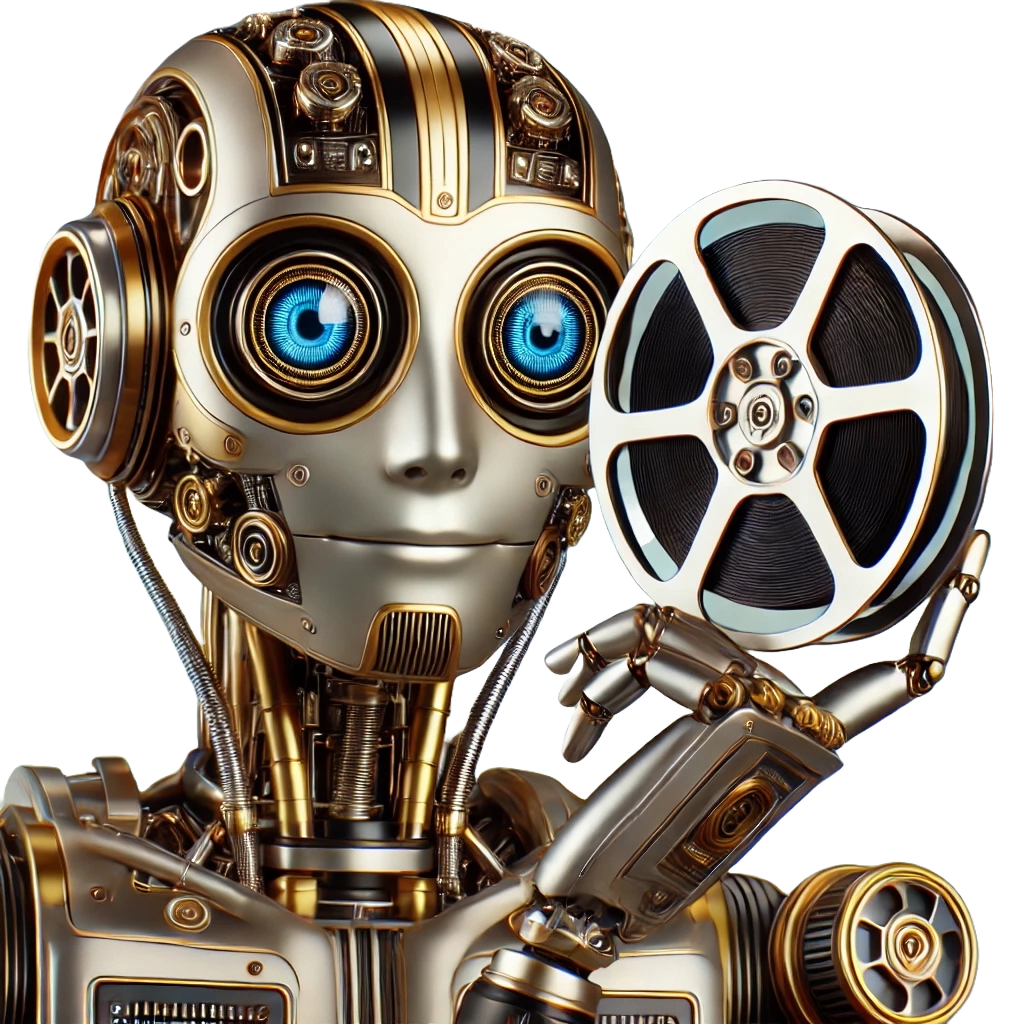}: A Large Multimodal Model for Pixel-Level \\ Visual Grounding in Videos}

\author{Shehan Munasinghe$^{1}$ \quad Hanan Gani$^{1}$ \quad  Wenqi Zhu$^{2}$ \quad Jiale Cao $^{2}$    \\ Eric Xing$^{1,3}$ \quad Fahad Shahbaz Khan$^{1,4}$ \quad  Salman Khan$^{1,5}$
\\[0.5em]
    {\fontsize{10.5pt}{12pt}\selectfont \textsuperscript{1}Mohamed bin Zayed University of AI, \textsuperscript{2}Tianjin University, 
    \textsuperscript{3}Carnegie Mellon University},\\ {\fontsize{10.5pt}{12pt}\selectfont  \textsuperscript{4}Linköping University,
    \textsuperscript{5}Australian National University
    }\\[0.0em]
    {\fontsize{9pt}{10pt}\selectfont shehan.munasinghe@mbzuai.ac.ae, hanan.ghani@mbzuai.ac.ae
    }\\[0.0em]
{\hypersetup{urlcolor=magenta}
  \fontsize{11pt}{12pt}\selectfont {\tt \href{https://mbzuai-oryx.github.io/VideoGLaMM}
  {https://mbzuai-oryx.github.io/VideoGLaMM}}}
}

\begin{document}
\maketitle
\begin{abstract}
Fine-grained alignment between videos and text is challenging due to complex spatial and temporal dynamics in videos. Existing video-based Large Multimodal Models (LMMs) handle basic conversations but struggle with precise pixel-level grounding in videos. To address this, we introduce VideoGLaMM, a LMM designed for fine-grained pixel-level grounding in videos based on user-provided textual inputs. Our design seamlessly connects three key components: a Large Language Model, a dual vision encoder that emphasizes both spatial and temporal details, and a spatio-temporal decoder for accurate mask generation. This connection is facilitated via tunable  V$\rightarrow$L and L$\rightarrow$V adapters that enable close Vision-Language (VL) alignment. The architecture is trained to synchronize both spatial and temporal elements of video content with textual instructions. To enable fine-grained grounding, we curate a multimodal dataset featuring detailed visually-grounded conversations using a semiautomatic annotation pipeline, resulting in a diverse set of 38k video-QA triplets along with 83k objects and 671k masks. We evaluate VideoGLaMM on three challenging tasks: Grounded Conversation Generation, Visual Grounding, and Referring Video Segmentation. Experimental results show that our model consistently outperforms existing approaches across all three tasks. 
\end{abstract}    
\section{Introduction}
\begin{figure}[t]
    \centering
    \includegraphics[width=0.48\textwidth]{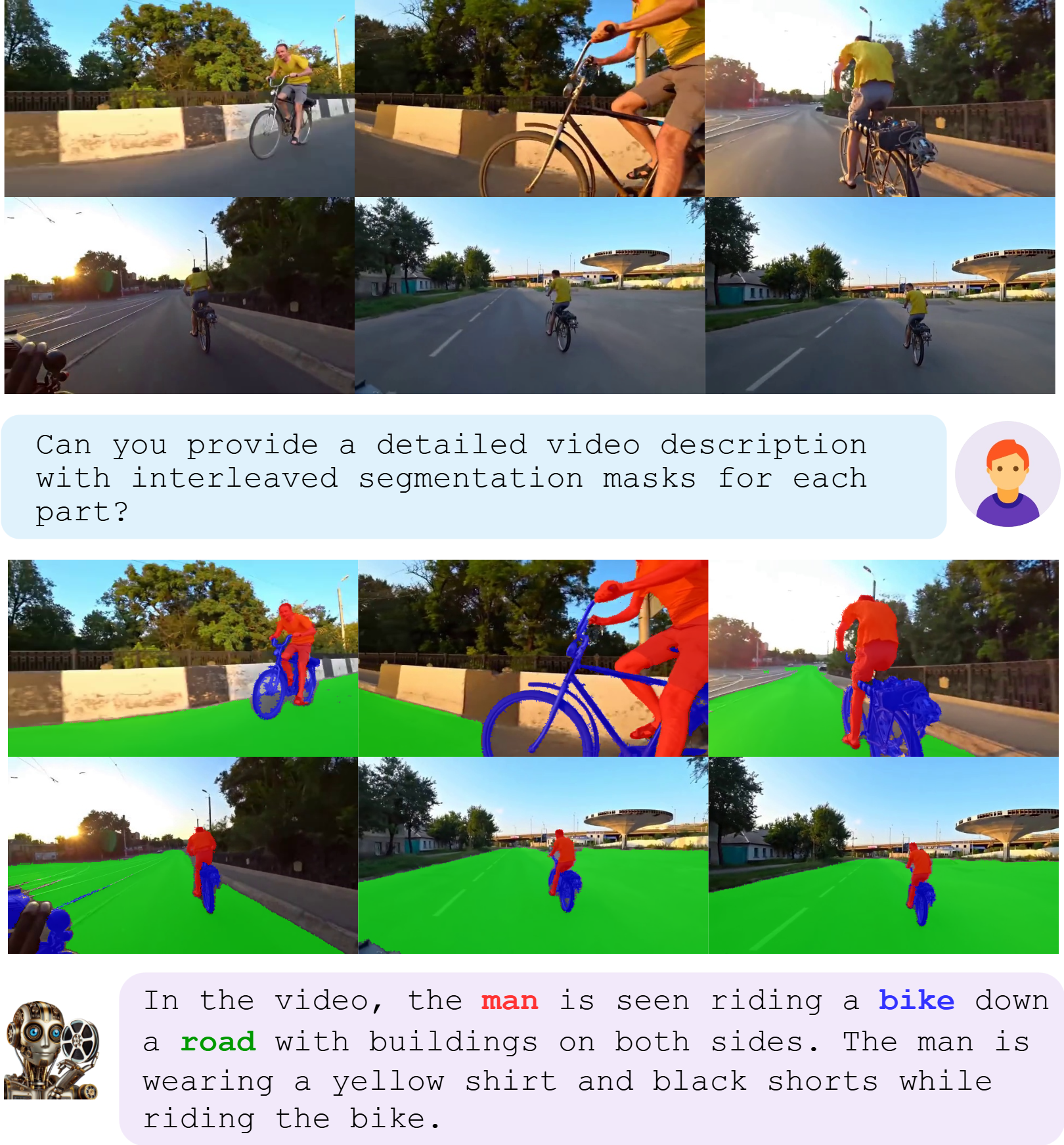}
    \vspace{-1.7em}
    \caption{\textbf{Grounded Conversation with VideoGLaMM. }
Our proposed multimodal video conversational model provides text responses grounded at the pixel level in the input video. The generated masks are spatio-temporally consistent across frames. The fine-grained grounded outputs from VideoGLaMM describe different levels of granularity, e.g., person, objects (bike), stuff (road), and explain object and scene attributes. Existing Video-LMMs do not
offer pixel-level grounded conversational capability. }
\label{fig:teaser}
\vspace{-1.5em}
\end{figure}
The rise of Large Language Models (LLMs) has significantly advanced progress in language-based tasks \citep{conf/nips/BrownMRSKDNSSAA20,vicuna2023,conf/acl/DuQLDQY022,conf/nips/Ouyang0JAWMZASR22,journals/corr/abs-2302-13971}. 
Their success in solving language-based complex reasoning tasks has led to their adoption in visual domains, resulting in Large Multimodal Models (LMMs). 
To align textual and visual modalities, previous works \citep{zhu2023minigpt,liu2024visual,conf/nips/Dai0LTZW0FH23,li2023blip,lai2023lisa} train a projection layer or a cross-attention block that maps visual features into the latent space of an LLM. 
This straightforward adaptation has enabled advanced spatial understanding, allowing detailed conversations about image content. Recently, these models have been extended to video, aligning textual instructions with the spatio-temporal inputs, leading to the development of Video-LMMs.

Existing Video-LMMs \citep{lin2023video,2023videochat,maaz2024videogpt+,damonlpsg2023videollama,Maaz2023VideoChatGPT,munasinghe2023pg,chen2023shikra}, similar to image-based LMMs, tune single or multiple projection layers to align videos with the language modality using the conventional visual instruction tuning paradigm. 
Although this simple alignment aids in understanding the global content of videos, it poses challenges in capturing localized object-specific context.
Consequently, 
existing works \citep{munasinghe2023pg,2023videochat,damonlpsg2023videollama,lin2023video} have demonstrated capabilities in video comprehension and dialogue,
they lack the crucial feature of fine-grained visual grounding, which aims to associate the LMM's response
to specific objects within the video input.
The ability of an LMM to generate visually grounded responses ensures that the model understands fine-grained spatial and temporal details in a video and can relate them with the generated text.

To bridge this gap, we introduce VideoGLaMM, a large video multimodal model capable of pixel-level spatio-temporal grounding. The model responds to natural language queries from the user and intertwines spatio-temporal object masks in its generated textual responses to provide a detailed understanding of video content. 
VideoGLaMM seamlessly connects three key components: a Large Language Model (LLM); dual vision encoders; and a spatio-temporal pixel decoder.
The dual vision encoders extract spatial and temporal features separately, which are jointly passed to the LLM to output responses rich in both spatial and temporal cues. 
Our spatio-temporal pixel decoder outputs the fine-grained object masks corresponding to the specific objects in the LLM output to visually ground its responses. 
These components are integrated via tunable Vision-to-Language (V→L) and Language-to-Vision (L→V) adapters that enable close vision-language alignment, unlike existing works that perform alignment with a single adapter.

As there currently exists no instruction-tuning dataset with fine-grained masks associated with video conversations, we present a benchmark instruction tuning dataset curated through a semi-automatic pipeline (Sec. \ref{sec:semiautomatic-pipeline}).
The dataset consists of 38k grounded video-QA triplet pairs with 83k objects and 671k fine-grained masks. The proposed benchmark dataset enables spatio-temporal modeling and significantly augments the capacity of the model to understand videos comprehensively, leading to state-of-the-art performance in grounded conversation generation, temporal grounding, and referring video segmentation
tasks under zero-shot settings. 

In summary, our contributions are as follows:
\begin{itemize}
    \item We introduce VideoGLaMM, a video large multimodal model, capable of pixel-level spatio-temporal grounding, featuring an end-to-end alignment mechanism.
    \item To achieve fine-grained spatio-temporal alignment, we introduce a benchmark instruction tuning dataset consisting of 38k grounded video-QA triplet pairs and 83k objects and roughly 671k fine-grained spatio-temporal masks.
    \item We assess the performance of VideoGLaMM across diverse tasks spanning grounded conversation generation (GCG), visual grounding, and referring video segmentation, where it achieves state-of-the-art performance.
\end{itemize}

\section{Related work}
\textbf{Large Multi-modal Models (LMMs).} Vision-language models like \citep{radford2021learning} have made notable advancements, demonstrating impressive zero-shot capabilities using millions of noisy image-text pairs during training. These models have been effective in various applications, from detection and segmentation \citep{bangalath2022bridging,liang2023open} to more complex tasks such as 3D understanding and video analysis \citep{ni2022expanding,liu2023internchat,wang2021actionclip,rozenberszki2022language}. The rise of LLMs has driven significant progress in Natural Language Processing (NLP) tasks and sparked interest in developing LMMs. Early models 
\citep{alayrac2022flamingo,awadalla2023openflamingo} incorporate visual information into intermediate embeddings for a frozen LLM using a cross-attention mechanism, trained on billions of image-text pairs to align visual and linguistic modalities. Similarly, BLIP-2 \citep{li2023blip} introduces Q-Former to better align visual features with language space. MiniGPT-4 \citep{zhu2023minigpt} and LLAVA \citep{liu2024visual} finetune on detailed image descriptions using a single projection layer to align a frozen visual encoder with a frozen LLM. Subsequent LLaVA series models \citep{liu2024improved} employ a multi-layer perceptron and a two-stage instruction tuning to refine the alignment process. While these works work on static images, our work focuses on efficiently aligning videos with linguistic cues.

\noindent
\textbf{Video LMMs.}
Recent advancements in image-based multimodal models have paved the way for video LMMs, which are essential for handling spatiotemporal sequences. Models such as VideoChat \citep{2023videochat}, Video-LLaMA, Video-ChatGPT \citep{damonlpsg2023videollama}, Video-LLAVA \citep{lin2023video} and Video-GPT+ \citep{maaz2024videogpt+}  extend the capabilities of LLMs to video domain by aligning video features with language, followed by instruction tuning on datasets annotated by either GPT models or humans. While these models have shown effectiveness in video comprehension, they still face limitations in fine-grained spatio-temporal modeling and visual grounding. This restricts their ability to accurately understand or localize specific objects and detailed segments within videos, highlighting the need for further advancements in developing better multimodal models capable of visual grounding.

\begin{figure*}[ht]
    \centering
    \includegraphics[width=\textwidth]{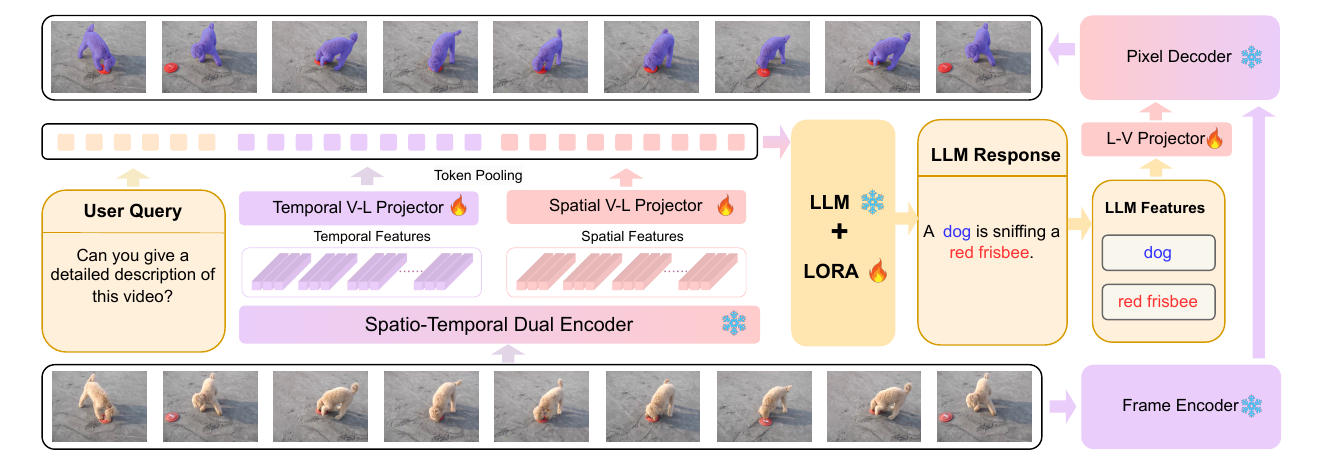}
    \caption{\textbf{Working of VideoGLaMM}. VideoGLaMM consists of a dual spatio-temporal encoder for encoding image and video level features. The spatial features represent the local information and the temporal features represent global information. The spatial and temporal tokens are passed through V-L adapters and concatenated with the text tokens, before feeding to LLM. A L-V projector is employed to align LLM's response with the visual space of pixel decoder. Finally, the aligned LLM features along with the frame features from a frame encoder are passed to a grounded pixel decoder, to obtain the fine-grained object masks corresponding to the LLM response. }
    \label{fig:method}
\end{figure*}

\noindent
\textbf{Visual Grounding. }
Recently Grounded LMMs \citep{peng2023kosmos,rasheed2023glamm,chen2023shikra,lai2023lisa,zhao2023bubogpt,wang2023all,you2023ferret} have made significant strides in enhancing visual and language comprehension and excel in complex localization tasks. These models demonstrate proficiency in tasks such as referring expression comprehension and image segmentation, highlighting the advanced image understanding capabilities of LLMs.
Approaches such as \cite{peng2023kosmos,chen2023shikra,wang2023all} primarily focus on creating a language-based context for visual grounding. In contrast, \cite{zhao2023bubogpt} integrates visual elements with language, while \cite{lai2023lisa} leverages vision-language embeddings to produce segmentation masks. Additionally, \cite{rasheed2023glamm} is adept at generating natural language responses linked with object segmentation masks, facilitating detailed visual-textual interactions. However, these models are limited to image-based applications and do not extend to video understanding.
Recently, \cite{munasinghe2023pg} incorporates audio transcripts
alongside visual and textual data for a more detailed video understanding. 
However, it combines pre-trained modules that cannot be trained end-to-end, which results in lack of fine-grained spatiotemporal modeling.
Similarly, \cite{bai2024one} introduced a new video grounding model, but their architecture employs only a spatial encoder-decoder setup and does not address the GCG task.
To this end, we propose VideoGLaMM, which leverages a novel fine-grained alignment strategy to align language instruction across both spatial and temporal dimensions, facilitating more finegrained video understanding.

\section{VideoGLaMM}

\subsection{Overview}
In this work, we introduce VideoGLaMM, a multi-modal video LMM with spatio-temporal pixel grounding capability. The task of spatio-temporal visual grounding focuses on linking a model's response to a user-specific text query with particular objects and regions within a video, ensuring that both spatial (what’s happening in each frame) and temporal (how things change over time) details are accurately reflected in the generated output (Fig.~\ref{fig:teaser}). By grounding responses in specific objects and actions across frames, the model demonstrates an understanding of both the evolving and static elements in a video, enabling it to produce responses that align closely with the visual narrative.

Our proposed VideoGLaMM is designed to achieve effective spatio-temporal grounding due to its ability to process spatial and temporal features simultaneously.
VideoGLaMM’s architecture (Fig.~\ref{fig:method}) leverages a dual-encoder structure: one encoder focuses on extracting spatial details from images, while the other captures temporal information from video sequences, ensuring complementary representation from both modalities. The visual features from both encoders are then integrated with a LLM using separate spatial and temporal adapters (V→L), guided by specific textual instructions.
The LLM outputs are aligned back with the visual space using an L→V adapter and further processed by a pixel decoder, which also takes video frames as input to produce the final grounded outputs.

For end-to-end spatio-temporal alignment, we train VideoGLaMM on our proposed fine-grained benchmark dataset. During training, we finetune the LoRA parameters of the LLM, along with V→L and L→V adapters. This approach seamlessly combines spatial and temporal data through an improved alignment mechanism and a precise grounding framework, enhancing the model’s capability for visual grounding and understanding. 

\subsection{Architecture}
The overall architecture of our VideoGLaMM consists of following components: (i) Spatio-Temporal Dual Encoder, (ii) Dual Alignment V-L Adapters, (iii) Large Language Model (LLM), (iv) Pixel Decoder. Below we provide a detailed description and working of each of component.

\noindent
\textbf{Spatio-Temporal Dual Encoder. }Our architecture consists of separate image and video encoders for extracting spatial and temporal features, thus leveraging the complementary strengths of both. This enables the model to have both local and global properties. The
image encoder $\mathcal{F}_g$, processes the $T$ video frames separately such that the input video $V \in \mathbb{R}^{T\times H \times W \times C}$. The output of the image encoder, represented by $f_g$, produces local spatial features that provide
frame-level context.
\begin{equation}
     f_{g} = \mathcal{F}_{g}(V), \quad V \in \mathbb{R}^{T\times H \times W \times C}
\end{equation}
Meanwhile, for extracting video features, we use segment-wise Sampling following \citep{maaz2024videogpt+} to obtain fine-grained temporal cues. Given an input video $V \in \mathbb{R}^{T\times H \times W \times C}$, we divide it into $K$ segments, where each segment consists of $s = \frac{T}{K}$ frames. The video encoder $\mathcal{F}_h$, operates on low-resolution video segments $V_{k} \in \mathbb{R}^{s\times H \times W \times C}$ yielding global features that provide segment-wise temporal context.
\begin{equation}
     f_{h} = \mathcal{F}_{h}(V_{k}), \quad V_{k} \in \mathbb{R}^{s\times H \times W \times C}
\end{equation}

\noindent
\textbf{Dual Alignment (V→L) Adapters}
To align visual features with the LLM space, we use two separate V→L adapters for image and video encoders. $\mathcal{W}_g$ represents the spatial adapter, and $\mathcal{W}_h$ represents the temporal adapter. These adapters project the visual features into the LLM's projection space, thus aligning the two modalities. The spatial and visual features corresponding to image and video samples after projecting from $W_g$ and $W_h$ are represented by $Z_g$ and $Z_h$, respectively.
\begin{equation}
    \begin{aligned}
        Z_{g} &= \mathcal{W}_{g}(f_g), \quad and \quad Z_{h} &= \mathcal{W}_{h}(f_h)
    \end{aligned}
\end{equation}

\noindent
\textbf{Large Language Model}
The tokenized spatio-temporal visual features are then concatenated with the textual tokens $Z_{text} \in \mathbb{R}^{L\times D_{t}}$ to obtain final feature embedding $\mathcal{Z} = [Z_{g}, Z_{h}, Z_{text}]$ which is fed into the \textbf{LLM}.
Thus, input to the \textbf{LLM} contains both the spatial and temporal cues for robust video understanding. We further expand the original \textbf{LLM} vocabulary
with a new token, i.e., \texttt{<SEG>}, which signifies the request
for the segmentation output. Thus the \textbf{LLM} response $\textbf{E}$ can be described as,
\begin{equation}
    \textbf{E} = \textbf{LLM}(\mathcal{Z}) = \textbf{LLM}(\textbf{[}Z_{g}, Z_{h}, Z_{text}\textbf{]})
\end{equation}
The LLM output $\textbf{E}$ contains the \texttt{<SEG>} whenever the task requires to generate the segmentation mask.

\vspace{0.5em}
\noindent
\textbf{Pixel Decoder} Our Pixel decoder consists of a prompt encoder ($\mathcal{H}$) and a mask decoder $\mathcal{D}$, capable of predicting masks with spatio-temporal grounding. The pixel decoder is adapted to videos and can implicitly process temporal information. 
The last layer embeddings from the LLM denoted as $\textit{l}_\texttt{seg}$ corresponding to \texttt{<SEG>} token is extracted, which is enriched with both spatial and temporal cues. The LLM embeddings act as prompts for the mask decoder and are processed by the prompt encoder. Simultaneously, we extract visual features of the input frames $V$ using a grounded frame encoder $\mathcal{P}$ which is aligned with pixel decoder and is further equipped with the ability to produce multi-scale features during training.
For aligning the output embeddings from LLM with the pixel decoder, we train an (L→V) adapter layer $\mathcal{W}_p$ between the LLM and prompt encoder such that the output from the adapter is denoted as $\textbf{e}^{p}_\texttt{seg} = \mathcal{W}_p(\textit{l}_\texttt{seg})$. The $\textbf{e}^{p}_\texttt{seg}$ is fed to prompt encoder $\mathcal{H}$, such that the encoded output $\mathcal{H}(\textbf{e}^{p}_\texttt{seg})$ is used to prompt the mask decoder. The encoded prompts $\mathcal{H}(\textbf{e}^{p}_\texttt{seg})$ along with the grounded visual features $\mathcal{P}(V)$ are passed to mask decoder $\mathcal{D}$. Subsequently, $\mathcal{D}$ produces the output mask $\textbf{M}$. 
\begin{equation}
    \textbf{M} = \mathcal{D} \Bigl(\mathcal{P}(V),\mathcal{H}(\textbf{e}^{p}_\texttt{seg})\Bigl)
\end{equation}



\begin{figure*}[!ht]
    \centering
    \includegraphics[width=1.0\textwidth]{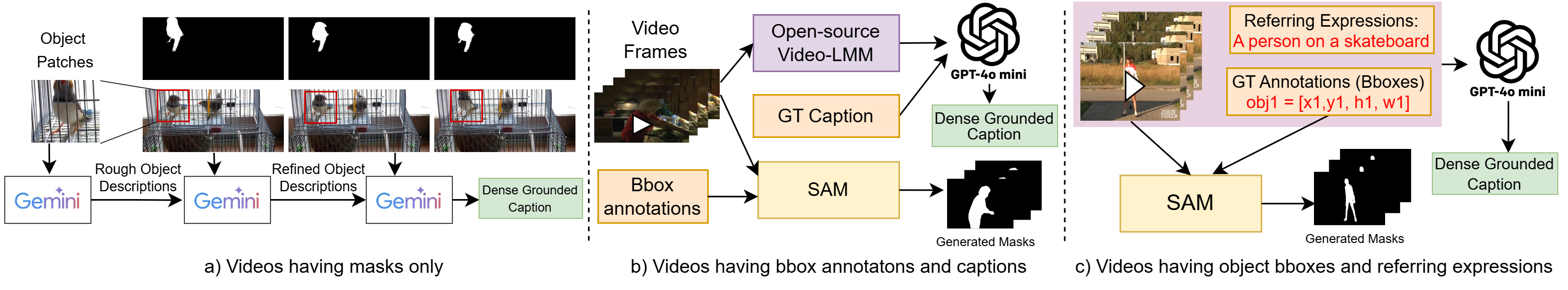}
    \caption{\textbf{Proposed Semi-automatic Annotation Pipeline}. Our dataset for grounded conversation generation (GCG) is built from three video dataset types: i) \textit{Videos having masks only:} Object patches are extracted from video frames using masks and processed by the Gemini-Pro model for initial object descriptions, which are then refined to produce detailed object captions. These refined captions and masks are again fed to Gemini-Pro model to create dense grounded captions. ii) \textit{Videos having bbox annotations and captions:} Frames are first processed with a Video-LMM to generate a comprehensive caption which is combined with the original caption and fed to GPT-4o to obtain dense grounded captions. Masks are generated using frames and ground-truth bounding boxes with the SAM model. iii) \textit{Videos having object bboxes and referring expressions:} Frames, bounding boxes, and referring expressions are input to GPT-4o for dense grounded captions, while masks are generated by feeding frames and bounding boxes to the SAM model.}
    \label{fig:annotation-method}
    \vspace{-1em}
\end{figure*}
\vspace{-1em}

\subsection{Training Strategy}
\label{Training-Strategy}
 We train VideoGLaMM end-to-end in a single stage. As stated above, we use a dual encoder consisting of separate image and video encoders for processing spatial and temporal inputs to obtain local and global features, respectively. These encoders are initialized with weights of strong pre-trained encoders. During training, we keep the encoders fixed and only train the V→L adapters $\mathcal{W}_g$ and $\mathcal{W}_h$ associated with these encoders. These adapters are used to project the spatio-temporal visual features in the space of LLM and align the two modules. The spatio-temporal encoder is kept frozen and only the V→L adapters are updated. The textual features from the last layer of LLM, rich in spatial and temporal cues, are projected into the space of the pixel decoder using a multi-layer projection L→V adapter $\mathcal{W}_p$. For the LLM, we keep its weights frozen and only finetune LoRA \citep{hu2021lora} parameters during training. Both the frame encoder and pixel decoder are instantiated with pre-trained weights 
 We keep the frame encoder and pixel decoder frozen and only train the L→V adapter layer. We optimize the output of the LLM by minimizing the cross entropy $\textbf{CE}$ objective between the autoregressively obtained text output and dense grounded ground-truth caption. For the output of mask decoder, we optimize the intersection over union (IOU) between the predictions of mask decoder and ground-truth masks denoted as $\mathcal{L}_{masked}$. The total loss is the sum of  $\textbf{CE}$ loss and masked loss.
 \begin{equation}
    \mathcal{L}_{total} = \textbf{CE} + \mathcal{L}_{masked}
\end{equation}
The first component of $\mathcal{L}_{total}$ ensures that the LLM generates textual embeddings that not only align with the ground truth but also offer informative spatio-temporal cues to the mask decoder for effective grounding. The second component facilitates efficient grounding by leveraging these textual cues from the LLM.

\section{Our Benchmark \& Annotation Pipeline}
\label{sec:semiautomatic-pipeline}
Our benchmark video dataset comes from different sources: YTVIS \cite{vmt}, BURST \cite{athar2023burst} ActivityNet  entities \citep{zhou2019grounded}, Refer-YTVOS \citep{seo2020urvos}, MeViS \cite{MeViS}, VidSTG \cite{zhang2020does} and HCSTVG \cite{tang2021human}. 
To create fine-grained grounded captions, we develop a semi-automated pipeline (Fig. \ref{fig:annotation-method}) that ensures high-quality and scalable annotation. Our annotation pipeline is categorized into three streams based on the availability of the ground truth annotations. We explain each stream below. 

\noindent\textbf{a) Videos with only Mask annotations:}
Fig. \ref{fig:annotation-method}(a) shows the annotation process for the videos having only masks as ground truth labels. To generate the corresponding dense grounded caption, we use following steps: \textbf{i)} \textit{Object Description Generation:} For each object in the video, we begin by creating a bounding box based on the ground truth mask provided in the annotation file. This bounding box allows us to crop the object from each frame, producing a sequence of image patches that capture the object throughout the video. We then feed these image patches to the Gemini-Pro model \cite{reid2024gemini} to obtain a rough description of each object in the video. 
\textbf{ii)}  \textit{Object Description Refinement:} The bounding boxes from the previous stage are superimposed on the corresponding video frames, and the entire video is then fed into the Gemini-Pro model to obtain a more accurate and detailed description of the objects. 
\textbf{iii)} \textit{Caption Generation:} The bounding boxes of corresponding objects overlayed across the video frames are labeled according to their object IDs. Then, we input these frames into the Gemini-Pro model to obtain dense captions. This results in a comprehensive description of the video. Finally, we manually review the \{obj\_id\} in the generated video captions based on the video content. \textbf{iv)} \textit{Detailed Dense Captions.}
To enhance the detail and accuracy of the video captions, we leverage two advanced Video LMMs: Video-LLAVA \citep{damonlpsg2023videollama} and LLAVA-NeXT \citep{liu2024llavanext}. Using the semi-automatically generated captions as a reference, we integrate and refine the outputs from these models, merging their results to produce the final, comprehensive dense captions. 

\noindent\textbf{b) Videos with Bounding Box annotations and Captions:}
Fig. \ref{fig:annotation-method}(b) shows the annotation process for the videos having both captions and object bounding box (Bbox) annotations. To obtain the corresponding dense grounded caption, the video frames are first passed to an open-source Video-LMM \cite{liu2024llavanext} to obtain a detailed caption, which is fed along with the reference ground truth caption to GPT-4o mini \cite{GPT4V} to obtain the final dense grounded caption. The Bbox annotations are used as a prompts to SAM model \cite{kirillov2023segment} which takes the video frames as input and provides the masks corresponding to the objects.

\noindent\textbf{c) Videos with Bounding Box annotations and Referring Expressions:}
Fig. \ref{fig:annotation-method}(c) shows the annotation process for the videos having object bounding box (Bbox) annotations and referring expressions corresponding to different objects. The video frames along with Referring expressions and Bbox annotations are prompted to GPT-4o mini, which provides the corresponding dense grounded caption. To obtain the masks corresponding to the objects, the video frames are fed to SAM model, which is prompted with Bbox annotations of the objects.

Overall, our proposed GCG dataset has 38,788 grounded video-QA triplets along with 83,877 objects and 6,71,016 fine-grained masks in total. 
We further curate a separate test set of 308 refined video-QA triplets with 826 objects and 22762 finegrained masks for grounded conversation generation evaluation task. 
\section{Experimental Setup}

\begin{figure*}[!t]
    \centering
    \includegraphics[width=\textwidth]{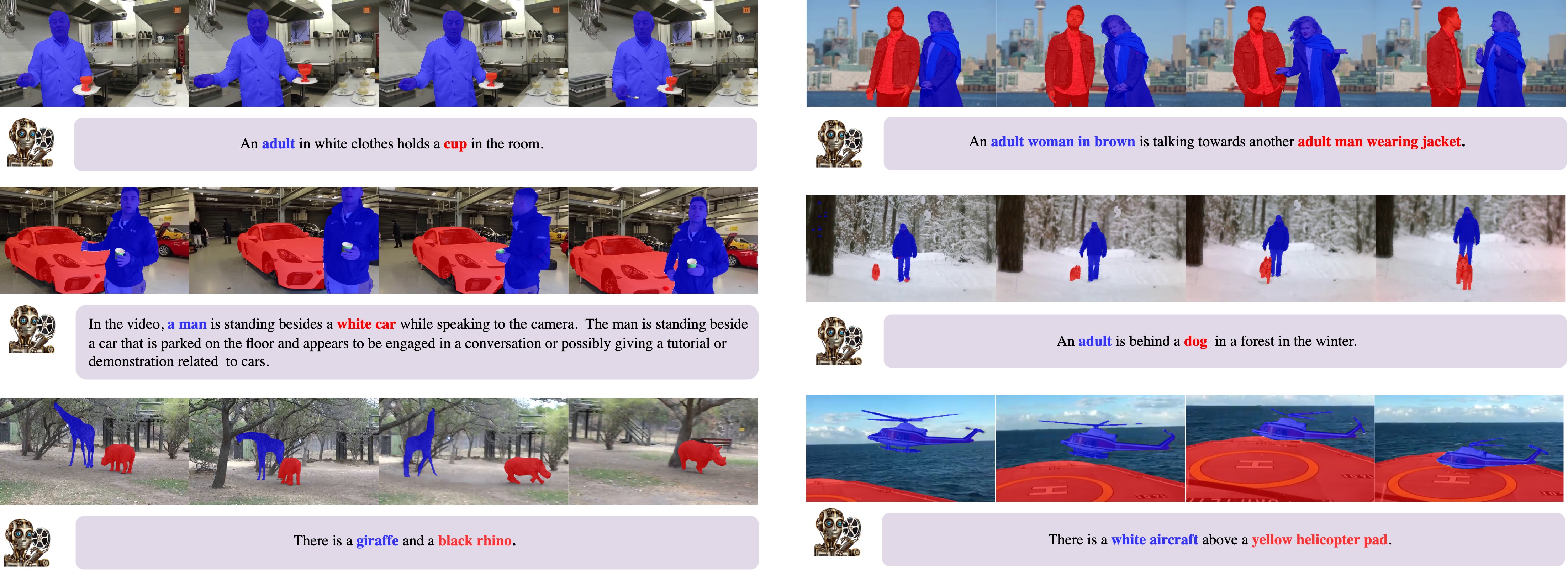}
    \caption{\textbf{Qualitative results of VideoGLaMM on grounded conversation generation (GCG)}. Given user queries, the VideoGLaMM generates textual
responses and grounds objects and phrases using pixel-level masks, showing its detailed understanding of the video. }
    \label{fig:qualitative}
    \vspace{-1em}
\end{figure*}

\noindent
\textbf{Implementation details. }
Our spatio-temporal dual encoders follow the design of image and video encoders from \cite{maaz2024videogpt+}. For the image encoder, we use a pretrained CLIP ViT-L/14 (336 × 336)\cite{radford2021learning} model, and for the temporal encoder, we select the pretrained encoder of InternVideov2 (224 × 224) \citep{wang2024internvideo2}. The V→L projectors are initialized with the weights of MLP adapter from \cite{maaz2024videogpt+}.
The LLM is instantiated with Phi3-Mini-3.8B \cite{abdin2024phi} weights. 
Both the frame encoder and pixel decoder are initialized with SAM2 \citep{ravi2024sam2} encoder-decoder weights. 
The training (Sec. \ref{Training-Strategy}) is carried out end-to-end on 4 Nvidia A100 40GB GPUs with a distributed training based on DeepSpeed \cite{deepspeed}.

\noindent
\textbf{Datasets. }
We train the model on our proposed grounded conversation (GCG) dataset containing 38k grounded video-QA triplets along with 83k objects and 671k fine-grained masks. 
During training, we also include a variety of other image and video segmentation datasets with our proposed benchmark dataset for more robust alignment. Our choice of image segmentation datasets include: ADE20K \citep{zhou2017scene}, COCO-Stuff \citep{caesar2018coco}, LVIS-PACO \citep{ramanathan2023paco}, refCOCO, refCOCO+, refCLEF, refCOCOg \citep{kazemzadeh-etal-2014-referitgame}, LLaVA-Instruct-150k \citep{liu2023visual}, ReasonSeg \citep{lai2023lisa} and GranDf \citep{rasheed2023glamm}. For video egmentation datasets we include train samples from Refer-DAVIS17 \citep{Khoreva_arxiv2018} and VideoInstruct100K \citep{Maaz2023VideoChatGPT}.

\noindent
\textbf{Tasks.}
We evaluate VideoGLaMM on three challenging tasks: grounded conversation generation (GCG), visual grounding, and referring video segmentation. For grounded conversation generation, we curate a separate dataset of 308 refined video-QA triplets containing 826 objects and 22,762 fine-grained masks, following our proposed annotation pipeline. 
For Visual Grounding, we evaluate our model on challenging VidSTG \citep{zhang2020does} dataset, considering only the interrogative sentences as done by \cite{munasinghe2023pg}.
In the case of motion-guided video object segmentation, we leverage the MeViS \citep{MeViS} validation dataset. All the results on MeViS dataset are obtained via official CodaLab evaluation suite. We also report referring video object segmentation results on additional Ref-DAVIS-17 \cite{khoreva2019video} dataset.

\noindent
\textbf{Evaluation metrics. }
For GCG task, we use mean Intersection over Union (mIOU) and Recall to determine the correctness of generated masks, and METEOR, CIDEr and CLAIR score for determining the goodness of conversational output. In the case of visual grounding, we report mean Intersection over Union (mIOU) to quantify the performance. Finally, for referring video segmentation, We report Region Jaccard $\mathcal{J}$, Boundary F measure $\mathcal{F}$, and their mean $\mathcal{J\&F}$.

\noindent
\textbf{Baselines. }We compare our VideoGLaMM with two challenging baselines employing LLMs capable of visual grounding: PG-Video-LLaVA \cite{munasinghe2023pg} and GLaMM \cite{rasheed2023glamm}. Since GlaMM is designed for pixel grounding in images, we enable temporal properties in GLaMM by augmenting its architecture with SAM2. For referring segmentation, we also compare VideoGLaMM with the recently released Video-LISA \cite{bai2024one}.  

\noindent
\textbf{Training Recipe. }
 VideoGLaMM follows a gradual training schedule. We do not train VideoGLaMM on our GCG dataset directly from the start, rather we take a gradual approach. We first train the model on image and video segmentation datasets until epoch 20 and then introduce our GCG dataset and train the model until epoch 30. This training recipe ensures that model learns both the spatial and temporal cues effectively.

\subsection{Grounded Conversation Generation}
The Grounded Conversation Generation (GCG) task aims to provide video-level detailed captions with specific phrases directly tied to corresponding segmentation masks in the video frames.
For example, ``\texttt{<An adult> in white clothes holds a <cup> in the room}", as shown in the first row of  Fig. \ref{fig:qualitative}, features how each bracketed
phrase is anchored to a unique segmentation mask. 
This creates a densely annotated caption that aligns textual descriptions with visual regions in the frames, enriching the video’s contextual interpretation. 
To obtain GCG output, we query the model with the following sample prompt: “\texttt{Provide a detailed description of the image. Respond with interleaved segmentation masks for the corresponding parts of the answer.}” The model generates a detailed caption along
with interleaved segmentation masks, employing the format “\texttt{<p>An adult woman in brown</p><SEG> is talking to another <p>adult man wearing jacket</p><SEG>}” as shown in the third row of Fig.~\ref{fig:qualitative}. We use special tokens, namely \texttt{<p>,
</p> and <SEG>}, to delineate the start and end of each
phrase and its corresponding region mask, respectively.

 As shown in Table \ref{table:gcg}, our proposed Video-GLaMM performs better in generating detailed captions containing references to objects in the video frames, as is evident from high METEOR, CIDEr and CLAIR scores. Regarding the quality of masks, VideoGLaMM consistently outperforms baselines in terms of mIOU and Recall scores, signifying a higher overlap with ground-truth masks. Fig. \ref{fig:qualitative} further shows the qualitative visualizations of VideoGLaMM on GCG samples.

\begin{table}[!ht]
\centering
\resizebox{\columnwidth}{!}{%
\begin{tabular}{lccccc}
\toprule
\textbf{Model} & \textbf{mIoU} & \textbf{Recall} & \textbf{METEOR} & \textbf{CIDEr} & \textbf{CLAIR} \\
\midrule
PG-Video-LLaVA \cite{munasinghe2023pg} & 24.03 & 0.093 & 0.10 & 0.01 & 15.0 \\
GLaMM \cite{rasheed2023glamm} + SAM2 \cite{ravi2024sam2} & 28.60 & 0.117 & 0.097 & 0.15 & 22.9 \\
\rowcolor{violet!10} VideoGLaMM & \textbf{62.34} & \textbf{0.375} & \textbf{0.103} & \textbf{0.59} & \textbf{28.2} \\
\bottomrule
\end{tabular}%
}
\caption{\textbf{Evaluation on grounded conversation generation (GCG):} VideoGLaMM shows superior performance in generating accurate video-level captions which are tied to corresponding segmentation masks in the video frames.}
\label{table:gcg}
\vspace{-1em}
\end{table}

\subsection{Referring Video Segmentation}
\label{sec:Referring Video Segmentation}
For referring video segmentation, the output should be grounded as per the given phrase, pointing towards specific instances in the video. Given a sentence or referring expression containing a specific object instance, the goal is to localize the object instances present across the video frames. This task operates in an open vocabulary setting, assessing the model's ability to localize objects both spatially and temporally. Given a referring phrase expression \texttt{Phrase}, we prompt the model using the following instruction prompt to obtain the instance masks:  ``\texttt{What is \{Phrase\} in this video? Respond with segmentation masks}". Table \ref{table:ref-seg} shows results on challenging MeViS dataset for motion-guided referring video segmentation. Both the region Jaccard $\mathcal{J}$ and boundary F-measure $\mathcal{F}$ are high in the case of VideoGLaMM, significantly outperforming the baselines. Similarly, the mean  $\mathcal{J\&F}$ follows the same trend. Additionally, the scores corresponding to VideoLISA are reported with post-processing step. Notably, VideoLISA involves an additional post-processing step to boost performance. Therefore, we further fine-tune the VideoGLaMM on the task of referring segmentation post epoch 30 until epoch 40.  Clearly, VideoGLaMM outperforms the Video-LISA (post-processed) on both $\mathcal{J}$ and $\mathcal{F}$, including the mean $\mathcal{J\&F}$. Additionally, VideoGLaMM outperforms baselines on Ref-DAVIS-17 dataset. The improved performance of VideoGLaMM can be credited to its training pipeline, which seamlessly integrates spatio-temporal dynamics into the model.



\begin{table}[!th]
\centering
\setlength{\tabcolsep}{12pt}
\begin{minipage}{0.43\textwidth}
    \resizebox{\textwidth}{!}{
    \begin{tabular}{llccc}
    \toprule
    \textbf{Model} & $\mathcal{J}$ & $\mathcal{F}$ & $\mathcal{J\&F}$ \\ 
    \midrule
    LMPM (MeViS baseline) \cite{MeViS} &37.2 &34.2 &40.2   \\
    
    PG-Video-LLaVA \cite{munasinghe2023pg} & 18.35 & 19.39 & 18.87 \\
    GLaMM \cite{rasheed2023glamm}+ SAM2 \cite{ravi2024sam2} & 35.80 & 41.50 & 38.66 \\
    VideoLISA \cite{bai2024one} & 41.30 & 47.60 & 44.40 \\
    \hline
    \rowcolor{violet!10} VideoGLaMM & \textbf{42.07} & \textbf{48.23} & \textbf{45.15} \\
    \bottomrule
    \end{tabular}
    }
\end{minipage}%
\hspace{0.01\textwidth}
\begin{minipage}{0.46\textwidth}
\vspace{0.2em}
\caption{\textbf{Performance comparison of VideoGLaMM on MeViS:} VideoGLaMM shows superior performance on motion grounding and segmenting referring objects in the videos.  }
\label{table:ref-seg}
\end{minipage}
\vspace{-2em}
\end{table}

\begin{table}[!th]
\centering
\setlength{\tabcolsep}{18pt}
\begin{minipage}{0.44\textwidth}
    \resizebox{\textwidth}{!}{
    \begin{tabular}{llccc}
    \toprule
    \textbf{Model} & $\mathcal{J}$ & $\mathcal{F}$ & $\mathcal{J\&F}$ \\ 
    \midrule
    LISA-7B \cite{lai2023lisa} &61.9  &54.9 &58.4  \\
    LISA-13B \cite{lai2023lisa} &64.6  &56.8 &60.7   \\
    TrackGPT-7B \cite{zhu2023tracking} &67.0 &59.4 &63.2 \\
    TrackGPT-13B \cite{zhu2023tracking} &70.4 &62.7 &66.5  \\
    VideoLISA \cite{bai2024one} &72.7 &64.9 &68.8 \\
    \hline
    \rowcolor{violet!10} VideoGLaMM & \textbf{73.3} & \textbf{65.6} & \textbf{69.5} \\
    \bottomrule
    \end{tabular}
    }
\end{minipage}%
\hspace{0.01\textwidth}
\begin{minipage}{0.44\textwidth}
\vspace{0.5em}
\caption{\textbf{Performance comparison of VideoGLaMM on Ref-DAVIS-17:} VideoGLaMM shows superior performance on segmenting referring objects in the videos. }
\label{table:ref-seg-davis}
\end{minipage}
\vspace{-2em}
\end{table}

\subsection{Visual Grounding}
\begin{table}[!ht]
\vspace{-1em}
\centering
\begin{minipage}{0.45\textwidth}
    \resizebox{\textwidth}{!}{
    \begin{tabular}{lc}
    \toprule
    \textbf{Model} & \textbf{VidSTG (interrogative mIoU)} \\ 
    \midrule
    PG-Video-LLaVA-7B \cite{munasinghe2023pg} & 34.20 \\
    PG-Video-LLaVA-13B \cite{munasinghe2023pg} &  35.10 \\
    GLaMM \cite{rasheed2023glamm} + SAM2 \cite{ravi2024sam2} & 38.63 \\
    \rowcolor{violet!10} VideoGLaMM & \textbf{39.66} \\
    \bottomrule
    \end{tabular}
    }
\end{minipage}%
\hspace{0.01\textwidth}
\begin{minipage}{0.48\textwidth}
\vspace{0.5em}
\caption{\textbf{Performance comparison of VideoGLaMM with other models on spatial grounding:} Results on VidSTG (interrogative) benchmark highlights VideoGLaMM's superior ability in correlating textual instructions with the visual frames.}
\label{table:visual-grounding}
\end{minipage}
\vspace{-1.8em}
\end{table}
\noindent
To quantitatively assess VideoGLaMM’s visual grounding capability, we conduct quantitative evaluations on the
benchmark test set of VidSTG dataset. The visual grounding task measures the adeptness of the model at correlating textual descriptions with
visual elements in the video, a critical aspect of contextual comprehension. This ability is crucial for applications that integrate continuous visual data with language. The output of this task is refined masks that correlate with the given caption \texttt{\{caption\}}. To obtain the visual grounding output, we query the model with interrogative captions.
For these captions, the prompt format follows ``\texttt{\{caption\} Please respond with a segmentation masks.}".
Table \ref{table:visual-grounding} shows VideoGLaMM's improved visual grounding precision as it outperforms the baselines, demonstrating its fine-grained understanding. 

In addition to the above downstream tasks, in Sec. \ref{supp:additional-qualitative} of supplementary, we also integrate VideoGLaMM into a conditional video generation model \cite{Shi_2024_CVPR}. VideoGLaMM provides temporally coherent masks that guides generative model in editing videos effectively. Please refer to Sections \ref{supp:additional-quantitative} and \ref{supp:additional-qualitative} of supplementary for additional quantitative and qualitative results respectively.

\subsection{Ablation studies}

\noindent
\textbf{Effect of Spatio-Temporal Dual Encoder.} We employ separate image and video encoders to process spatial and temporal information. While spatial processing induces local information, temporal processing helps learn global features. Both are necessary from the perspective of grounding. To verify the effectiveness of dual spatio-temporal encoder, we conduct an ablation study to measure the effectiveness of each encoder for grounded conversation generation (GCG) task (see Table \ref{ablation:encoder}). We notice that using only an image encoder gives suboptimal results, as we notice a drop in both the localization and captioning metrics. Using only video branch leads to the highest mIOU; however, relatively lower METEOR, CIDEr, and CLAIR scores. To obtain an optimal mIOU and good conversational abilities,  VideoGLaMM uses both image and video encoders. 

\begin{table}[!ht]
\centering
\resizebox{\columnwidth}{!}{%
\begin{tabular}{lccccc}
\toprule
\textbf{Encoder Configuration} & \textbf{mIoU} & \textbf{Recall} & \textbf{METEOR} & \textbf{CIDEr} & \textbf{CLAIR} \\
\midrule
Image encoder &60.06	&0.395	&0.081	&0.371	&18.9 \\
Video encoder &64.62	&0.375	&0.097	&0.568	&26.5 \\
\rowcolor{violet!10} Dual encoder & 62.34 & 0.375 & 0.103 & 0.590 & 28.2 \\
\bottomrule
\end{tabular}%
}
\caption{\textbf{Effect of Spatio-Temporal Dual Encoder:} We obtain low performance using only spatial (image) encoder. Using only a video encoder gives the highest mIOU but lower scores on CLAIR, METEOR and CIDEr. For a better trade-off, we employ dual (image and video) encoders to have accurate, grounded conversations.}
\label{ablation:encoder}
\vspace{-1em}
\end{table}

\noindent
\textbf{Spatial vs Spatio-temporal Pixel decoder.} Pixel decoder in VideoGLaMM can operate in two configurations. The first configuration processes video frames individually, ignoring temporal consistency. The second configuration employs both spatial and temporal branches for spatio-temporal context. Table \ref{ablation:decoder} demonstrates the impact of spatiotemporal decoder on the GCG task. Results indicate that using only the spatial configuration reduces performance, with a nearly 3\% drop in mIOU scores compared to the spatio-temporal configuration. Similarly, metrics like METEOR, CIDEr, and CLAIR also show a decline, underscoring the importance of using spatio-temporal configuration for pixel decoder. 

\begin{table}[!ht]
\centering
\resizebox{\columnwidth}{!}{%
\begin{tabular}{lccccc}
\toprule
\textbf{Decoder Configuration} & \textbf{mIoU} & \textbf{Recall} & \textbf{METEOR} & \textbf{CIDEr} & \textbf{CLAIR} \\
\midrule
Spatial decoder &59.68	&0.369	&0.097	&0.553	&26.7\\
\rowcolor{violet!10} Spatio-temporal decoder &62.34	&0.375	&0.103	&0.59	&28.2 \\
\bottomrule
\end{tabular}%
}
\caption{\textbf{Spatial vs Spatio-temporal Pixel decoder:} We observe that using Pixel decoder without the temporal branch gives limited performance as the model faces difficulties in temporal grounding. When using temporal branch, the performance on both the temporal grounding and grounded LLM response improves indicating the importance of temporal processing in VideoGLaMM.}
\label{ablation:decoder}
\vspace{-1.5em}
\end{table}

\noindent
\textbf{Effect of number of frames for Pixel Decoder.} The pixel decoder receives the raw input frames encoded via frame encoder as input for predicting fine-grained grounded masks. During training, the pixel decoder also receives ground-truth masks which act as supervision signals.
To provide more temporal supervision, we feed the pixel decoder with multiple input frames to enhance its temporal understanding. This allows it to learn semantic information that generalizes across frames. Table \ref{ablation:supervision-frames} shows the performance when 4 and 8 frames are input to the decoder. We observe that while the mIOU with 8 frames is slightly lower compared to 4 frames, the conversational quality measured by METEOR and CLAIR is higher. Hence, to achieve a decent mIOU with higher conversational output, we stick to 8 frames in the paper.

\begin{table}[!ht]
\centering
\resizebox{\columnwidth}{!}{%
\begin{tabular}{lccccc}
\toprule
\textbf{Decoder Input frames} & \textbf{mIoU} & \textbf{Recall} & \textbf{METEOR} & \textbf{CIDEr} & \textbf{CLAIR} \\
\midrule
4 frames &63.82	&0.37	&0.094	&0.659	&27.2\\
\rowcolor{violet!10} 8 frames &62.34	&0.37	&0.103	&0.590	&28.2 \\
\bottomrule
\end{tabular}%
}
\caption{\textbf{Effect of number of frames for Pixel Decoder:}  We observe that using 4 supervision frames for pixel decoder gives better mIOU but relatively modest conversation quality measured by METEOR and CLAIR. With 8 supervision frames, mIOU slightly decreases while the conversational quality increases. 
}
\label{ablation:supervision-frames}
\vspace{-0.8em}
\end{table}






\noindent
\textbf{Limitations and Future Work:}
Our GCG dataset plays a key role in enhancing the model’s grounding capabilities. While we validated annotations manually, some noise may still be present. Also, each scene contains several objects and the video descriptions do not exhaustively cover all objects in the scenes. A higher-quality densely annotated set could further boost model performance but would require substantial annotation resources. Additionally, VideoGLaMM struggles with objects of varying granularities, likely due to limited representation in the training data. Another improvement is to extend VideoGLaMM for longer videos, as the current GCG dataset mainly focuses on short-medium duration clips. 


\section{Conclusion}
We introduce VideoGLaMM, a LMM specifically designed to address the challenge of fine-grained pixel-level grounding in videos. By integrating a dual vision encoder with a spatio-temporal decoder and employing tunable Vision-Language adapters, our model achieves precise alignment between video content and textual instructions. To facilitate this alignment, we introduce a refined instruction-tuning dataset curated via a semi-automatic annotation pipeline. Our experimental evaluations across Grounded Conversation Generation, Visual Grounding, and Referring Video Segmentation tasks demonstrate that VideoGLaMM consistently outperforms existing models. 

{
    \small
    \bibliographystyle{ieeenat_fullname}
    \bibliography{main}
}
\clearpage
\appendix
\setcounter{page}{1}
\maketitlesupplementary

We provide  supplementary material for further understanding of certain sections of the main paper. We have arranged the sections as follows:
\begin{itemize}
    \item Additional Quantitative Results
    \item Additional Qualitative Results
     \item Instruction templates for Annotation pipeline
    \item Ethics and Societal Impact
\end{itemize}

\section{Additional Quantitative Results}
\label{supp:additional-quantitative}
We perform additional quantitative evaluations of VideoGLaMM on the Refer-YouTube-VOS dataset for the task of referring video object segmentation. 
As discussed in the Sec. \ref{sec:Referring Video Segmentation} of the main paper, for referring video segmentation, the output should be
grounded as per the given phrase, pointing towards specific
instances in the video. Table \ref{table:ref-seg-yvos} shows the results on Refer-Youtube-VOS dataset. Our VideoGLaMM outperforms all the contemporary best performing baselines, suggesting the enhanced grounding and localization capability of our model.

\begin{table}[!th]
\centering
\setlength{\tabcolsep}{18pt}
\begin{minipage}{0.44\textwidth}
    \resizebox{\textwidth}{!}{
    \begin{tabular}{llccc}
    \toprule
    \textbf{Model} & $\mathcal{J}$ & $\mathcal{F}$ & $\mathcal{J\&F}$ \\ 
    \midrule
    LISA-7B \cite{lai2023lisa} &50.6 &49.7 &50.2  \\
    LISA-13B \cite{lai2023lisa} &53.0 &52.1 &52.6  \\
    TrackGPT-7B \cite{zhu2023tracking} &57.4 &55.3 &56.4 \\
    TrackGPT-13B \cite{zhu2023tracking} &60.8 &58.1 &59.5 \\
    VideoLISA \cite{bai2024one} &65.7 &61.7 &63.7 \\
    \hline
    \rowcolor{violet!10} VideoGLaMM & \textbf{65.4} & \textbf{68.2} & \textbf{66.8} \\
    \bottomrule
    \end{tabular}
    }
\end{minipage}%
\hspace{0.01\textwidth}
\begin{minipage}{0.44\textwidth}
\vspace{0.5em}
\caption{\textbf{Performance comparison of VideoGLaMM on Refer-Youtube-VOS:} VideoGLaMM shows superior performance on segmenting referring objects in the videos. }
\label{table:ref-seg-yvos}
\end{minipage}
\end{table}

\section{Additional Qualitative Results}
\label{supp:additional-qualitative}
\subsection{Qualitative results on GCG and referring object segmentation}

\begin{figure*}[!h]
    \centering
    \includegraphics[width=0.8\textwidth]{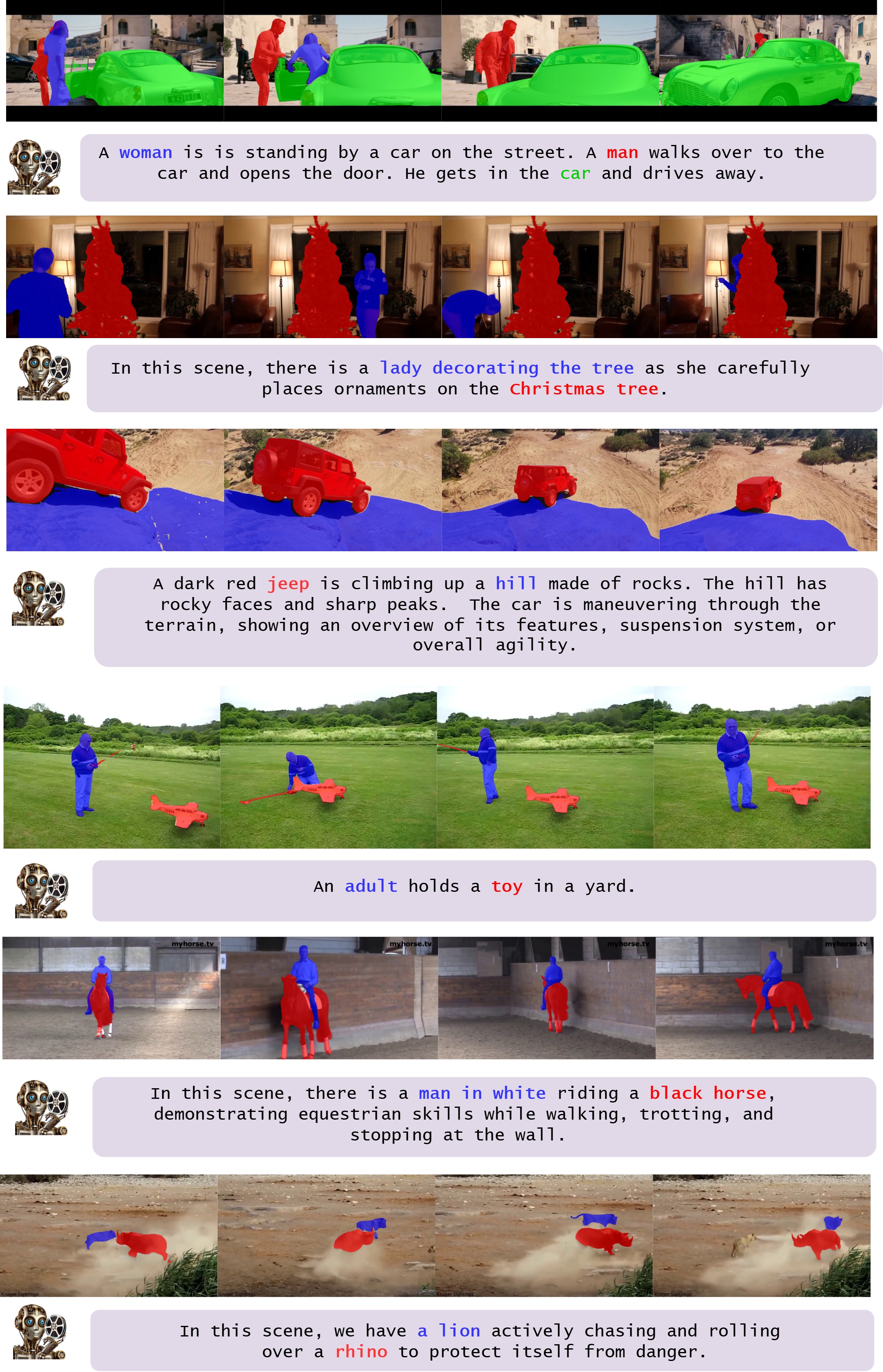}
    \caption{\textbf{Qualitative results of VideoGLaMM on GCG samples}. }
    \label{fig:supp_gcg}
\end{figure*}

\begin{figure*}[!h]
    \centering
    \includegraphics[width=0.85\textwidth]{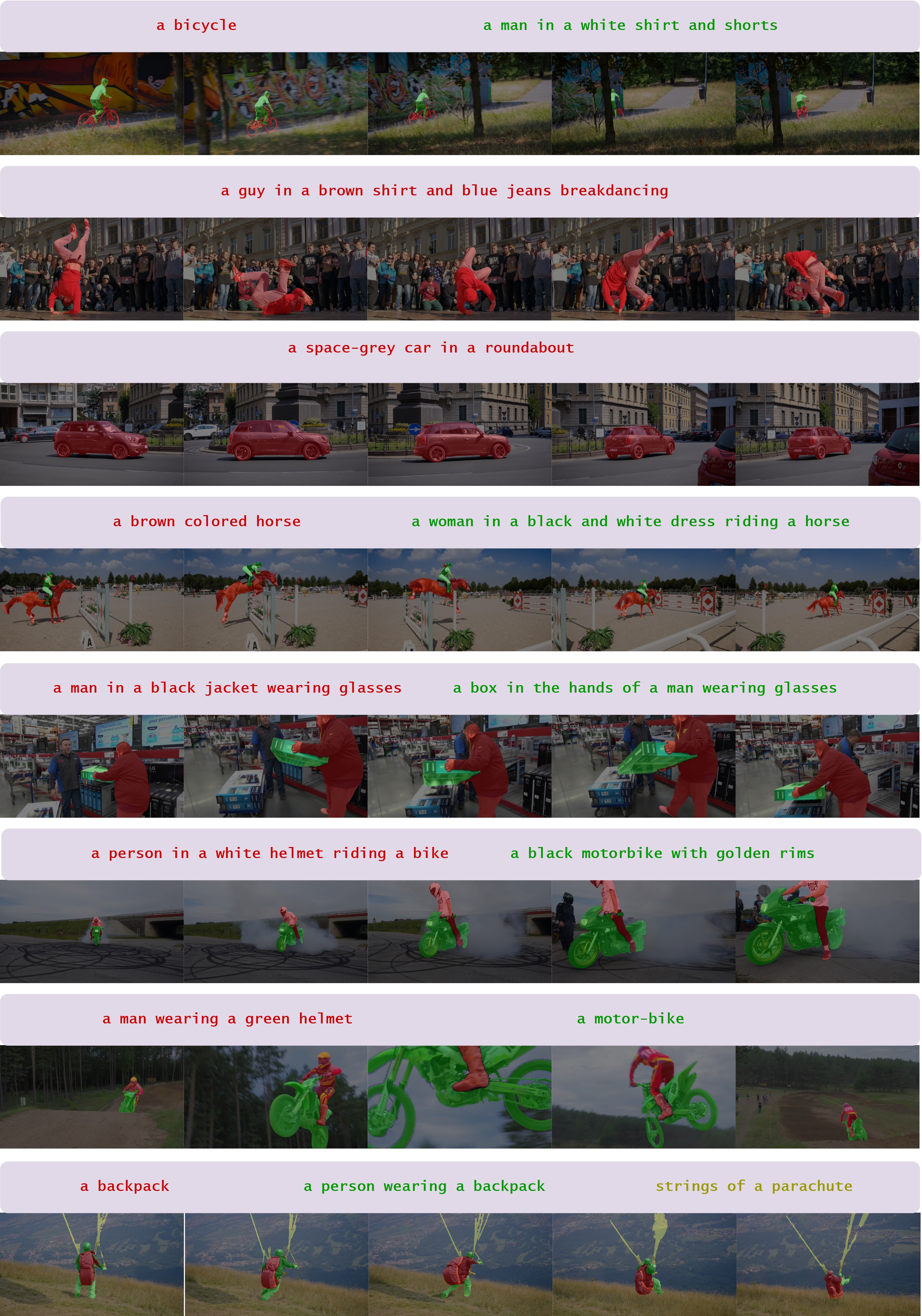}
    \caption{\textbf{Qualitative results of VideoGLaMM on referring object segmentation}. }
    \label{fig:supp_vos}
\end{figure*}

Figure~\ref{fig:supp_gcg} and Figure~\ref{fig:supp_vos} shows our model's performance on the GCG and referring object segmentation tasks respectively.

\subsection{Conditional Video Generation}
\begin{figure*}[ht]
    \centering
    \includegraphics[width=\textwidth]{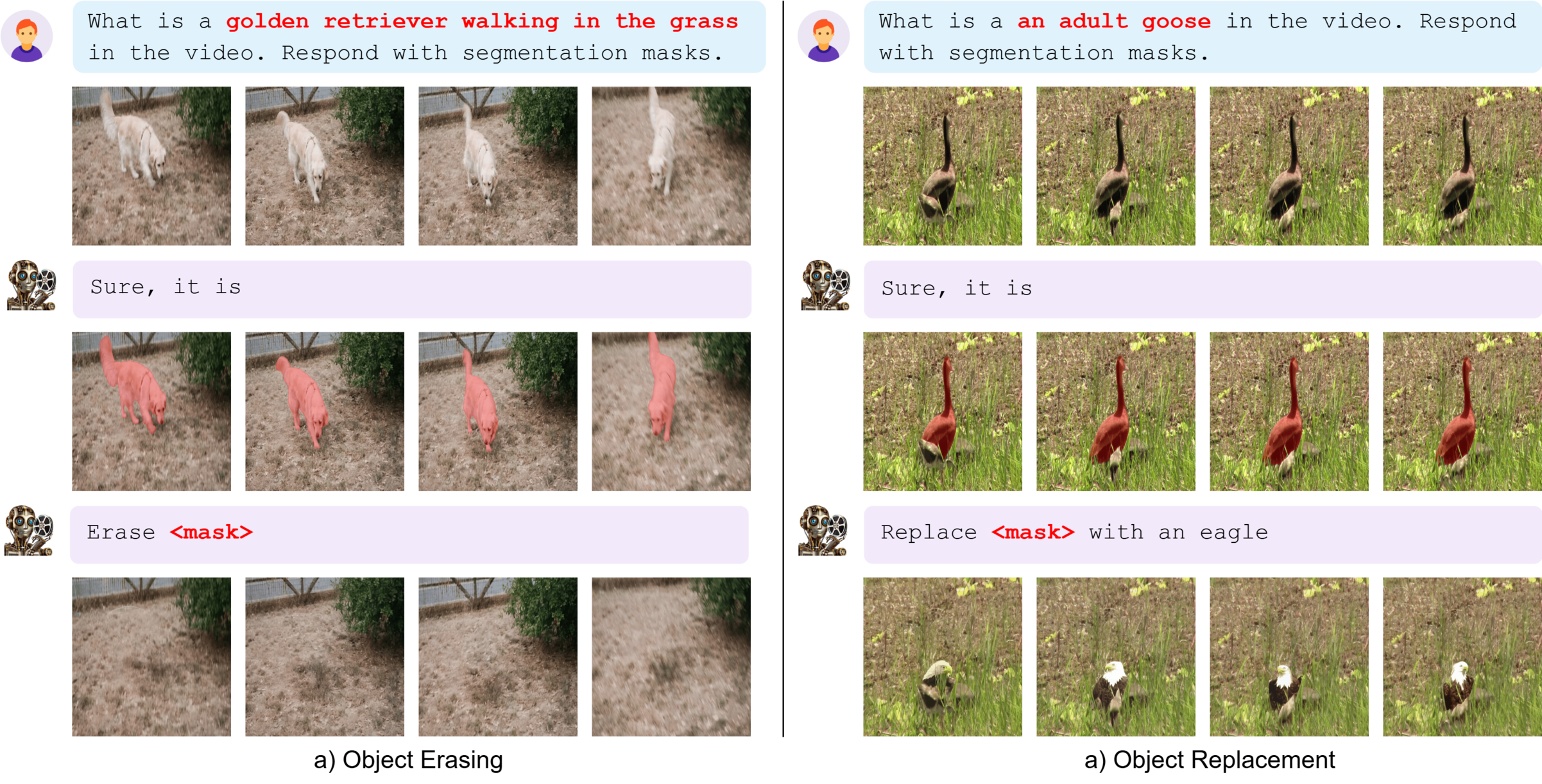}
    \caption{\textbf{Conditional Video Generation using VideoGLaMM}. }
    \label{fig:conditional_generation}
\end{figure*}

We incorporate VideoGLaMM into a conditional video generation model \cite{Shi_2024_CVPR}, designed to take an input video and modify it based on a specified condition (in this case, a mask) and a query prompt that outlines the desired edits. Given a video and a plain text query, VideoGLaMM identifies the relevant objects in the video by generating a precise mask for the object of interest. The generative model then utilizes the video, mask, and query prompt to produce the edited video. Figure \ref{fig:conditional_generation} illustrates two examples: object removal and object replacement, with the masks provided by VideoGLaMM. This demonstrates VideoGLaMM's adaptability in working with different models to address a wide range of video editing tasks.

\section{Instruction Prompt templates for Annotation pipeline}
\label{supp:instruct-templates}
We provide the instruction prompt templates used at each stage our semi-automatic pipeline below.

\vspace{1em}
\noindent
\textbf{a) Videos with only masks. }

\noindent
\rule{\linewidth}{0.1pt}

\noindent
i) Prompt for generating object patch descriptions

\noindent
\rule{\linewidth}{0.1pt}

\noindent
\texttt{These are frames from a video that I want to upload. What does the \{class\_name\} look like, and what is the \{class\_name\} doing?}

\noindent
\rule{\linewidth}{0.1pt}

\noindent
ii) Prompt for generating refined object descriptions

\noindent
\rule{\linewidth}{0.1pt}

\noindent
\texttt{These are frames from a video that I want to upload. Please refine this caption: \{caption from step 1\}. The instance in the video is highlighted by a rectangular box with the color corresponding to ID \{object\_id\}}

\noindent
\rule{\linewidth}{0.1pt}

\noindent
iii) Prompt for generating caption

\noindent
\rule{\linewidth}{0.1pt}

\noindent
\texttt{These are frames from a video that I want to upload. In the video, the ID number of the box is on the top left of the box. There are some instance captions:[The obj\_\{object\_id\} must be surrounded by a rectangular box with color number \{object\_id\}. It is a \{class\_name\}.\{object\_id’s caption from step 2\}, The obj\_\{object\_id\} must be surrounded by a rectangular box with color number \{object\_id\}. It is a \{class\_name\}.\{object\_id’s caption from step 2\}...] Generate a dense caption that describes the video in detail based on the video and instance captions, including all of the instances mentioned in the instance captions and other instances in the video. Ensure that each instance mentioned in the instance caption appears exactly once in the dense caption, followed by the format \{obj\_\} to indicate which instance caption the mentioned instance corresponds to. The \{obj\_\} must directly follow the noun representing the instance}

\noindent
\rule{\linewidth}{0.1pt}

\vspace{1em}
\noindent
\textbf{b) Videos having Bbox annotations and captions }

\noindent
\rule{\linewidth}{0.1pt}

\noindent
\texttt{Your task is to process video captions to make them more detailed and explanatory.
You are given a ground truth caption and a set of dense noisy captions.
Ground truth caption contains a description of the objects visible in a video, with noun phrases of significant objects surrounded by <p> and </p> tags, followed by a [SEG:x] tag.\\
Dense noisy captions contain additional information about the video, but they may be redundant or less precise than the ground truth caption.\\
Your task is to paraphrase the ground truth caption by incorporating relevant information from the dense noisy captions.\\
The refined caption should be more detailed and explanatory than the ground truth caption.\\
The refined caption should preserve the original <p>, </p>, and [SEG:x] tags.\\
The refined caption should also preserve the identity of [SEG:x] tags, given by a unique identifier x.\\
\\
You may look at the following examples:\\
Example 1:\\
Ground truth caption:\\
A <p> weight </p> [SEG:1] lifter is in a <p> gym </p> [SEG:2] , and <p> he </p> [SEG:1] lifts a <p> barbell </p> [SEG:0]\\
Reference captions:\\
In the video, a man is lifting weights in a gym. He lifts the weights over his head and then drops them on the ground.\\
In the video, a person is seen lifting weights in a gym setting. The individual is focused on performing the weightlifting exercise, and their posture indicates a controlled and deliberate movement. The gym environment is equipped with various weightlifting equipment, and there are other people present in the background, suggesting a shared workout space. The person's attire and the equipment indicate that this is a dedicated space for physical fitness and strength training. The video captures a moment of physical exertion and dedication to fitness.\\
Output:\\
\{"refined\_caption": "In the video, <p> A man </p> [SEG:1] is lifting weights in a <p> gym </p> [SEG:2]. <p> He </p> [SEG:1] is lifting a <p> barbell </p> [SEG:0] over his head and then drops them on the ground."\}
\\
Example 2:\\
Ground truth caption:\\
The <p> man </p> [SEG:1] stands while holding onto the <p> swing </p> [SEG:0]\\
Reference captions: \\
In the video, a man is swinging on a swing set in a park. He is wearing a black shirt and is swinging back and forth while looking towards the camera.\\
In the video, a person is standing in a park, wearing a black shirt and dark pants. The individual appears to be posing or standing still, possibly enjoying the surroundings or waiting for someone. The park features a playground with visible equipment, such as a swing set, indicating a recreational area for children and families. The person is standing on a concrete surface, and there are trees and other greenery in the background, suggesting a peaceful and natural setting. The individual's pose and the environment create a calm and leisurely atmosphere.\\
Output:\\
\{"refined\_caption": "In the video, <p> a man </p> [SEG:1] is swinging on a <p> swing set </p> [SEG:0] in a park. He is wearing a black shirt and is swinging back and forth while looking towards the camera."\}\\
Example 3:\\
Ground truth caption:\\
<p> She </p> [SEG:1] puts shaving <p> cream </p> [SEG:2] on <p> her </p> [SEG:1] <p> leg </p> [SEG:0] and shaves <p> her </p> [SEG:1] <p> leg </p> [SEG:0] \\
Reference captions: \\
In the video, a person is seen sitting on a tub and shaving their legs with a razor.\\
In the video, a person is seen sitting in a bathtub, and their legs are being shaved with a razor. The individual appears to be focused on the shaving process, and there are no other significant actions or events occurring in the video. The person's posture and the position of the razor suggest a careful and deliberate approach to shaving their legs. The setting appears to be a private bathroom, and there are no other people or objects visible in the frame.\\
Output:\\
\{"refined\_caption": "In the video, <p> a woman </p> [SEG:1] is seen sitting in a bathtub, shaving <p> her </p> [SEG:1] <p> legs </p> [SEG:0] with a razor. <p> She </p> [SEG:1] is applying <p> shaving cream </p> [SEG:2] on <p> her </p> [SEG:1] <p> leg </p> [SEG:0]."\}\\
\\
Now please refine the following caption:\\
Ground truth caption:\\
\{gt\_caption\}\\
Reference captions:\\
\{reference\_captions\}\\
Please provide the refined caption in (JSON format, with a key refined\_caption.)}

\noindent
\rule{\linewidth}{0.1pt}

\vspace{1em}
\noindent
\textbf{c) Videos having Bbox annotations and referring expressions }

\noindent
\rule{\linewidth}{0.1pt}

\noindent
\texttt{Your task is to generate annotated video captions, given original unannotated video descriptions, the lists of subjects/objects in the video and the relation between them.
\\
For each video, you are given a relation between a subject and an object, along with the categories and target IDs of the subject and object.
Your task is to generate a new caption annotating the subject and object in the caption with the corresponding target IDs.
\\
\\
You may look at the following examples:
\\
Example 1: 
\\
Input :\\
  subject : \\
    target\_id :0, category : rabbit\\
  object :\\
    target\_id : 1, category : adult\\
  relation : lean\_on \\
  description: "there is a white rabbit leaning on an adult by the water".
  \\
Output:\\
{'caption': 'there is a [white rabbit](0) leaning on an [adult](1) by the water'}
\\
\\
Now please process the following.
\\
{video\_relation\_data}
\\
In the new caption, the noun phrases should be included within square brackets and object ID/IDs should be included within paranthesis. E.g. [noun phrase](object ID/IDs) .
\\
Please provide the generated caption in JSON format, with a key "caption".}

\noindent
\rule{\linewidth}{0.1pt}

\newpage
\newpage
\newpage
\section{Ethics and Societal Impact}
\label{supp:ethics}
Our proposed dataset utilizes video samples from existing datasets which are released under the open-source public license and do not pose any privacy concerns. To the best of our knowledge, the dataset does not portray any strong biases or discrimination. We urge for the responsible use of our dataset and model, promoting research progress while safeguarding
privacy.

\end{document}